\documentclass{article}

\PassOptionsToPackage{numbers, compress}{natbib}

\usepackage[final]{neurips_2020}

\usepackage[utf8]{inputenc} 
\usepackage[T1]{fontenc}    
\usepackage{amssymb}
\usepackage{amsthm}
\usepackage{natbib}
\usepackage{amsmath}
\usepackage{graphicx}
\usepackage{caption}
\usepackage{hyperref}       
\usepackage{url}            
\usepackage{booktabs}       
\usepackage{amsfonts}       
\usepackage{nicefrac}       
\usepackage{microtype}      

\usepackage{bm}
\usepackage[ruled,vlined]{algorithm2e}
\usepackage{float}

\newtheorem{theorem}{Theorem}
\newtheorem{prop}{Proposition}

\DeclareMathOperator{\ranks}{\operatorname{rank}_s}
\DeclareMathOperator{\rank}{\operatorname{rank}}

\usepackage{xcolor}

\title{Neural Abstract Reasoner}

\author{
Victor Kolev%
\thanks{Sofia High School of Mathematics, Bulgaria} \\
\texttt{victor.kolev@yahoo.com}
\And
Bogdan Georgiev%
\thanks{Fraunhofer IAIS, Research Center for ML (FZML) and Competence Center ML2R, Germany} \\
\texttt{bogdan.m.georgiev@gmail.com}
\And
Svetlin Penkov 
\thanks{Sciro Research, Bulgaria} \\
\texttt{svet@sciro.ai}
}

\begin{document}

\maketitle

\begin{abstract}
Abstract reasoning and logic inference are difficult problems for neural networks, yet essential to their applicability in highly structured domains. In this work we demonstrate that a well known technique such as spectral regularization can significantly boost the capabilities of a neural learner. We introduce the Neural Abstract Reasoner (NAR), a memory augmented architecture capable of learning and using abstract rules. We show that, when trained with spectral regularization, NAR achieves $78.8\%$ accuracy on the Abstraction and Reasoning Corpus, improving performance 4 times over the best known human hand-crafted symbolic solvers. We provide some intuition for the effects of spectral regularization in the domain of abstract reasoning based on theoretical generalization bounds and Solomonoff's theory of inductive inference.
\end{abstract}

\section{Introduction}
Extracting and reasoning with abstract concepts is a crucial ability for any learner that is to operate in combinatorially complex open worlds or domains with limited or structured data. It is well known that neural learners struggle to operate in such conditions due to their poor generalization capabilities in structured domains \cite{chollet2019measure, marcus2018deep}. In this work, we demonstrate that spectral regularization provides neural networks with a strong inductive bias towards learning and utilizing abstract concepts akin to a symbolic learner.

For that purpose, we employ the Abstraction and Reasoning Corpus (ARC) \cite{chollet2019measure} which contains tasks related to manipulating colored patterns in a grid. In order to successfully solve the tasks in the corpus an agent needs to be able to count, manipulate numbers, work with topological and geometric concepts as well as recognise the notion of objects. There are 400 training tasks and 400 (distinct) evaluation tasks. Each task has a small set of input-output example pairs (between 1 and 5) and a query input pattern. This is quite a challenging dataset due to the small amount of example data, large number of different tasks and their abstract nature.

So far, the best known solution, with a success rate of ~20\%, is the winner of the ARC 
Kaggle challenge, and it is a carefully hand-crafted symbolic solver written in approx. 
7k lines of C++ code. In this paper, we introduce the Neural Abstract Reasoner (NAR) which
achieves an accuracy rate of ~79\% and so outperforming even the best symbolic solver
created by a human. The NAR architecture contains a Differentiable Neural Computer
(DNC) that learns general problem solving skills and a Transformer network
responsible for solving particular task instances (see. Fig 2 in \citet{chollet2019measure}).
 7u
Importantly, spectral regularization plays a fundamental role in the successful training
of NAR. From a purely machine learning perspective, spectral regularization is known
to reduce the effective number of parameters in the network, however we provide some 
additional theoretical intuition and demonstrate that spectral regularization also
pushes the network towards finding algorithmically simpler solutions as 
recommended by Solomonoff's theory of inductive inference \cite{Vitanyi2008}.

\section{Related Work}

\paragraph{Neuro-symbolic architectures}
Hybrid neuro-symbolic approaches enable agents to solve structured tasks from raw data, while learning faster and being more robust to noise \cite{gaunt2017differentiable, verma2018programmatically, penkov2018learning, mao2018neuro}. However, the majority of 
methods proposed so far are designed with specific domains in mind, making them 
inapplicable to a broader range of tasks. A notable exception is the architecture
proposed by \citet{ellis2020dreamcoder}, which is capable of learning rules from geometry, vector algebra, and physics and solve tasks such as drawing pictures or building complete scenes. Importantly, these methods often require lots of data, which is in stark contrast with human capabilities.

The ARC dataset \cite{chollet2019measure} is specially designed to push research towards data efficient learners, as there are hunderds of tasks, each of which is represented by no more than 5 input/output examples. To the best of our knowledge, the Neural Abstract Reasoner, presented in this paper, is the first architecture that achieves a performance rate of ~79\% on the ARC dataset, outperforming state-of-the-art hand-coded symoblic systems by a factor of 4. The NAR architecture is a composition of a slowly learning Differentiable Neural Computer and a fast adapting Transformer network creating an outer learning and inner executing loops, as suggested in \cite{chollet2019measure}.

\paragraph{Complexity and generalization}
The analysis of complexity and generalization metrics applied to neural networks has formed a central line of theoretical ML research with a variety of recent breakthroughs in terms of PAC-based and compression methods (cf. \cite{Arora2018, Bartlett2017, Jiang*2020Fantastic, Suzuki2020, WeiMa2019} and the references therein). In particular, many generalization approaches based on spectral norm analysis have been so far proposed and investigated \cite{Neyshabur2017, Sanyal2020Stable}. However, to our knowledge the present work is the first to address the relationship between spectral norms' behaviour and abstract reasoning tasks, whereby a strong relationship between a classical spectral regularisation and the ability of a neural model towards learning abstract reasoning (concepts and rules) is demonstrated. As touched upon in Section \ref{sec:spec-reg}, one could draw motivation from well-known algorithmic information theory concepts such as Solomonoff inference and program generation based on least complexity \cite{Vitanyi2008, Blier2018, Schmidhuber1997}.

\section{Methods}
\paragraph{Description}
The ARC dataset consists of a train and evaluation portions $D$ and $D_v$, respectively. Each portion consists of 400 tasks (train tasks are augmented to $\approx$15000 through color permutations and rotations). The individual tasks $T$ are grouped in tags $\tau$ based on the skills needed to solve them \cite{tasktagging}. A task $T_j = \Big \{ [i^e_1, o^e_1]; \dots; [i^e_5, o^e_5]; [i^q, o^q] \Big \}$ consists of up to five example input-output pairs and one query pair. A neural learner has to infer a logical rule $\pi: i \rightarrow o$. All inputs and outputs are grids of variable sizes with 10 colors. A solution is correct only when all the pixels on the grid match.

First, we derive a latent representation of the grids with an InceptionNet-style \cite{szegedy2015inception} deterministic auto-encoder. Let $G \in \{0, \dots, 9\}^{10 \times 10}$ be a grid, and $\mathcal{E}: G \rightarrow \mathbb{R}^n, \mathcal{D}: \mathbb{R}^n \rightarrow G$ be the encoder and decoder, parametrized by $\theta$. We train an embedding network by minimizing the standard autoencoder cross-entropy loss $\min_\theta: L_e = \sum_{G \in D} H\big (G, \mathcal{D}(\mathcal{E}(G)) | \theta \big )$.

Next, we consider all latent grid embeddings, and we train a Differentiable Neural Computer \cite{graves2016hybrid} $\mathcal{M}_\mu$ with parameters $\mu$ to infer an instruction set $\psi$. All inputs are processed by a Transformer Decoder Stack $\mathcal{T}$ \cite{vaswani2017attention} with parameters $\eta$, which self-attends to all inputs and cross-attends to $\psi$:
$$ \psi = \mathcal{M}_\mu \Big ( \big \{ [i^e_1, o^e_1]; \dots; [i^e_5, o^e_5] \big \} \Big ), \hspace{1cm} \widehat{o}_j = \mathcal{T}_\eta \big ( i_j | \psi; i_{k, k \neq j} \big ). $$

The whole model is then trained end-to-end via ADAM \cite{kingma2014adam} to minimize the cross-entropy loss between the query target and the decoded test output prediction.
$$\min_{\mu, \eta}: L = \sum_{T \in D} H \Big ( \mathcal{D}(o^q), \mathcal{D}(\widehat{o}^q | \mu, \eta)  \Big )$$

We employ a two-stage curriculum during training, first training only a on a single tag $\tau$, and then expanding to the whole train dataset $D$. Additionally, during the first stage of training, spectral regularization \cite{yoshida2017spectral} with a larger $\lambda$ value is applied, which is then annealed in the second stage. When evaluating the model, we apply additional optimization steps (similar to \cite{finn2017model, krause2018dynamic}), as described in Algorithm~\ref{algo:eval}. See Appendix C for additional details.

\paragraph{Motivation}
We build on methods from \citet{santoro2016meta} and use a memory-augmented neural network (the Differentiable Neural Computer \cite{graves2016hybrid}) to derive context for the current task. The multiple read heads and attention mechanisms allow the DNC to relate the input and the output of a pair and compare them to the other input/output pairs that it has already processed. Unlike \citet{santoro2016meta}, we leverage a Transformer to carry out the task execution based on the DNC context. This decouples the learning of the instruction set from the program execution itself, and allows us to use the input/output relations directly, rather than the more standard $[i_t, o_{t-1}]$. Lastly, the Transformer network relates the inputs to each other, thereby exploiting similarities within them.

\paragraph{Performance}
 As this work is still in progress, these are preliminary results evaluated on grids up to $10 \times 10$. Nonetheless, we outperform all currently known solutions, including a hand-crafted symbolic solution (see Fig. \ref{fig:perf}). Spectral regularization proved instrumental for this, and other regularization methods did not yield any significant results (see Fig. \ref{fig:arc_results}).  

Without any additional adaptation steps \cite{finn2017model}, evaluation performance remains low at 1\%, while only after 3 steps, that number climbs up to 78.8\%. Analyzing more closely the network changes made by the adaptation steps, the gradient norm of $\mathcal{M}$ is $<1\times10^{-7}$, which implies that the DNC is acting as a true meta-learner, and only the Transformer requires a small change $[3\times 10^{-4},  7\times 10^{-3}]$ to execute the instructions flawlessly. We again attribute this generalization to spectral regularization.

\begin{minipage}{0.54\textwidth}

\begin{algorithm}[H]
	\SetAlgoLined
    Given: $D_v$ evaluation dataset;\\
    
    $\mathcal{E}, \mathcal{D}, \mathcal{M}_\mu, \mathcal{T}_\eta$; \\
    Hyperparameters: $\alpha$ step size; $k$ number of steps\\

	\For{\text{Task} $T \in D_v$}{
	    $\mu_0 \leftarrow \mu$, $\eta_0 \leftarrow \eta$\\

	    \For{step in 1:k}{
	        $\psi = \mathcal{M}_{\mu_{\text{step} - 1}} \Big ( \big \{ [i^e_1, o^e_1]; \dots; [i^e_5, o^e_5] \big \} \Big )$\\
	        $[\widehat{o}^e_1, \dots, \widehat{o}_e^5] = \mathcal{T}_{\eta_{step - 1}}  \big ( [i_1^e, \dots, i_5^e, i^q] | \psi \big ) $\\
	        Evaluate $\mathcal{L} = \sum_{i=1}^5 H \big ( \mathcal{D}(o_i^e), \mathcal{D}(\widehat{o}_i^e) \big )$ \\
	        Adjust parameters: \\
	        $\mu_{step} \leftarrow \mu_{step -1} - \alpha \nabla_{\mu_{step -1}} \mathcal{L}$ \\
	        $\eta_{step} \leftarrow \eta_{step - 1} - \alpha \nabla_{\eta_{step - 1}} \mathcal{L}$ \\
	    }

	    $\psi = \mathcal{M}_{\mu_k} \Big ( \big \{ [i^e_1, o^e_1]; \dots; [i^e_5, o^e_5] \big \} \Big )$\\
	        $[\widehat{o}^e_1, \dots, \widehat{o}_e^5, \widehat{o}^q] = \mathcal{T}_{\eta_k}  \big ( [i_1^e, \dots, i_5^e, i^q] | \psi \big ) $\\
	    Compute accuracy of $\widehat{o}^q$\\
	}

    \caption{NAR evaluation cycle}
    \label{algo:eval}
\end{algorithm}

\end{minipage}\hspace{0.03\textwidth}%
\begin{minipage}{0.43\textwidth}
    \centering
    \includegraphics[width=0.7\linewidth]{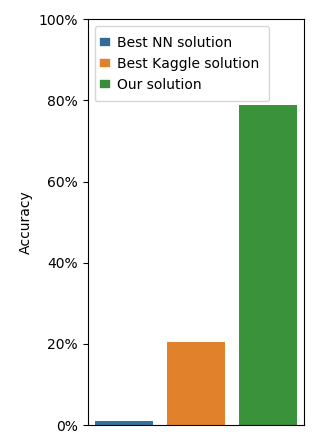}
    \captionof{figure}{We achieve 78.8\% evaluation accuracy on the ARC dataset. The reported results are calculated for 100 unseen tasks from the corpus. }
    \label{fig:perf}
\end{minipage}

\section{Effects of spectral regularization: stable ranks and complexity}
\label{sec:spec-reg}
The surprisingly significant effect of a simple spectral regularization strategy in the reasoning tasks suggests strong connections with generalization and the underlying model complexity estimates. On one hand, this motivates the analysis of spectral regularization in terms of some well-known generalization bounds (e.g. based on stable rank and spectral norms), however, we first discuss a perspective inspired by algorithmic information theory. Intuitively, abstract reasoning tasks are induced by a concise set of logic rules and combinatorial patterns, and, hence, it is natural to search for \textit{short} programs producing these rules - in this regard, we give motivation as to why spectral regularization naturally shrinks the search space towards \textit{shorter} programs.


\textbf{Spectral regularization and polynomials as algorithmically simple programs}. Classical methods from program inference and algorithmic information theory, such as Solomonoff inference and Occam's razor \cite{Vitanyi2008}, suggest that "simpler" program models are preferable in terms of forming abstract concepts and generalization - a formal approach towards such issues is given, e.g., by Kolmogorov complexity theory \cite{Vitanyi2008, Schmidhuber1997}. Although the evaluation of algorithmic complexity is a demanding task (Kolmogorov complexity is theoretically uncomputable), one could attempt to devise various proxy metrics that capture the algorithmic complexity of a given function/program. 

Here, in an attempt to evaluate and explain the algorithmic complexity of our models from a spectral-regularization perspective, we consider approximations in terms of a simple but flexible class $\pi_n$ of programs computing rational-coefficient polynomials of maximal degree $n \in \mathbb{N}$. Intuitively, approximating a model $f$ in terms of $\pi_n$ for lower values of $n$ corresponds to discovering programs of decreased algorithmic length (in terms of operations) that effectively compute $f$. In this direction, we bring forward some classical approximation theory results implying that lower spectral norms yield lower degrees of the approximation polynomial. To ease notation, here we discuss the 1-dimensional case, however, similar results hold for higher dimensions as well \cite{Lloyd2013}:

\begin{prop} \label{prop:Bernstein}
     Let $f:[0, 1] \rightarrow \mathbb{R}$ represent a model with Lipschitz constant $L$. Then, there exists a polynomial $B_n \in \pi_n$ of degree $n$, so that $\|f- B_n \|_{L^\infty[0, 1]} \leq \frac{3 L}{2 \sqrt{n}},$ where $\|.\|_{L^\infty[0, 1]}$ denotes the usual $\sup$-norm over the interval $[0, 1]$.
\end{prop}{}

Since the Lipschitz constant of a neural model is bounded above by the spectral norms of the layers $W_i$, spectral regularization gives control over $L$; moreover, a lower value of $L$ implies that one can decrease the polynomial degree $n$ and retain similar approximation qualities. These observations support our empirical results - introducing spectral regularization steers the model search space towards algorithmically simpler and more robust functions.



\textbf{Generalization via spectral norms and stable ranks}. We recall that the stable rank of a matrix $A$, $\ranks(A)$, is defined as the ratio $\|A\|^2_F / \|A\|^2_2$ and note that $\ranks(A)$ is at most the rank of $A$. The stable rank is intuitively understood as a continuous proxy to $\rank(A)$ and as a measure for the true parameter count of $A$. Now, let $f$ be a deep neural model consisting of $d$ layers whose corresponding weight matrices are denoted by $W_i, i=1, \dots, d$. Recent works (e.g. \cite{Neyshabur2017, Arora2018}) obtain generalization bounds on $f$, roughly speaking, in terms of the expression $\mathcal{O}\left(\prod_{i=1}^d \|W_i\|^2_2 \sum_{i=1}^d \ranks(W_i)\right),$ where $\|W_i\|_2$ denotes the spectral norm of the matrix $W_i$. A related stronger compression-based estimate in terms of so-called noise-cushions is obtained in \cite{Arora2018}. In other words, the generalization error is influenced by the spectral norms as well as stable ranks of the layers.

In this direction, we evaluated our model and the effect of stable ranks. Interpreting Fig. \ref{fig:sranks}, one observes that initially while the model adopts to the single task of $\texttt{pattern\_expansion}$ it increases and stabilizes a true parameter count $\sim 8000$; afterwards, the model is introduced to the full task bundle where a significant decrease of the stable ranks is observed - according to the last expression this leads to better generalization, and further implies that at the end of training one actually deals with simpler models with better compression properties.

\begin{minipage}{0.48\textwidth}
    \includegraphics[width=\linewidth]{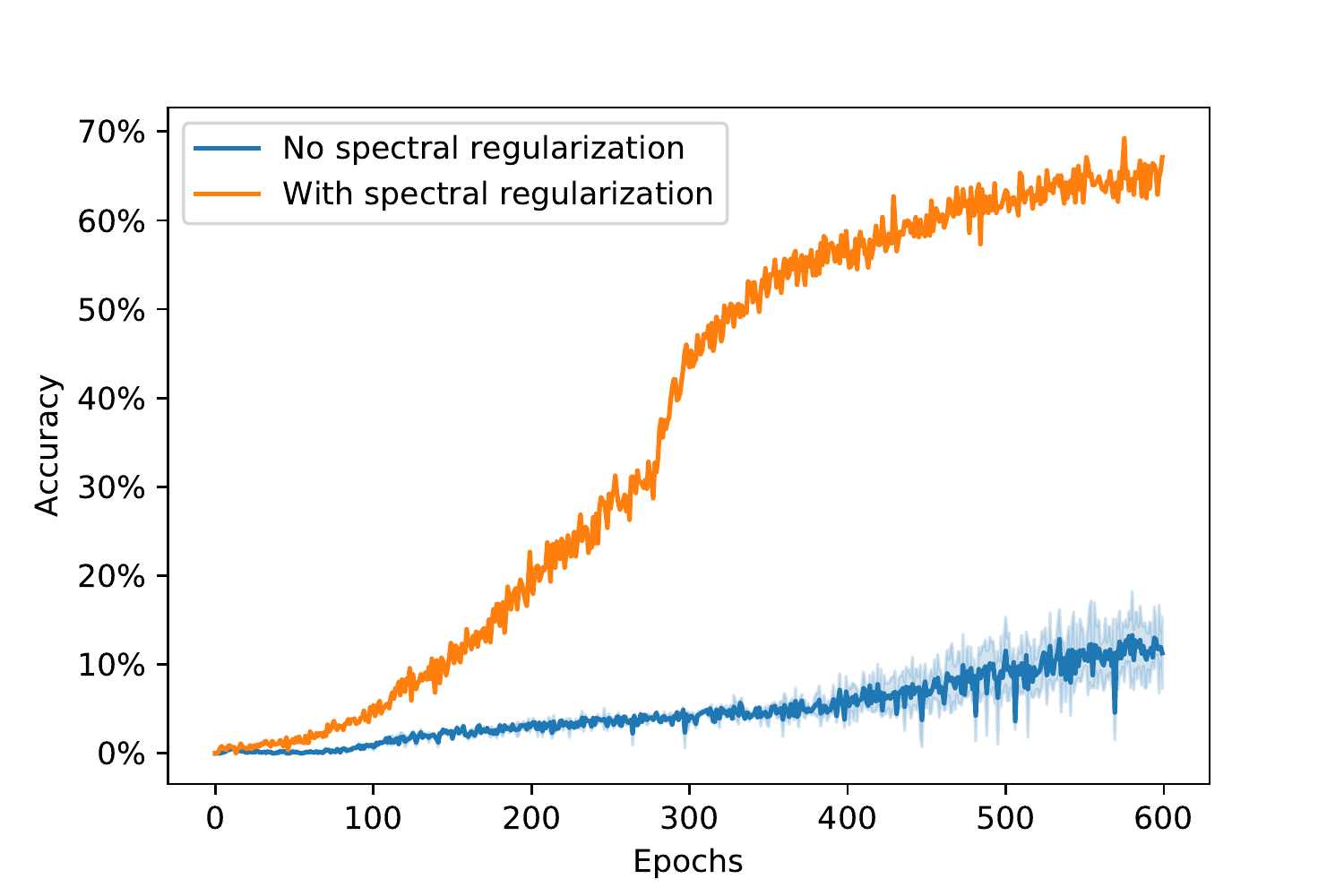}
    \captionof{figure}{Performance on the \texttt{pattern expansion} task from the ARC dataset. The exact same models (DNC Transformer) were trained with and without spectral regularization.}
    \label{fig:arc_results}
\end{minipage}\hspace{0.04\linewidth}%
\begin{minipage}{0.48\textwidth}
    \includegraphics[width=\linewidth]{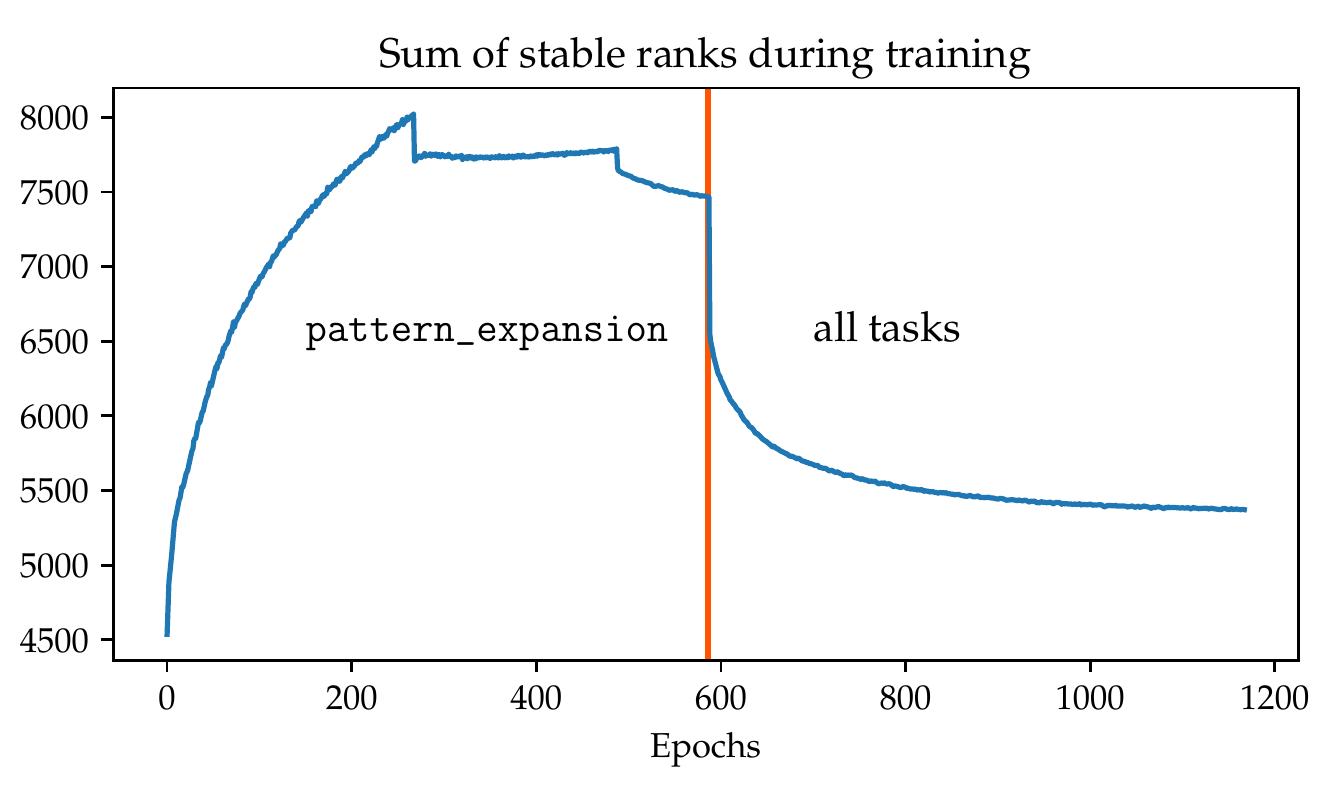}
    \captionof{figure}{Stable ranks vary drastically depending on the task distribution. With a larger pool of tasks, NN's stable ranks decrease rapidly, as it is optimized for greater generalization, as opposed to specialization to a particular task. }
    \label{fig:sranks}
\end{minipage}

\section{Conclusion and Acknowledgements}

We have demonstrated the spectral regularization provides neural learners with a significant boost in performance on abstract reasoning tasks. We believe that studying the complexity of the underlying models in the context of powerful frameworks such as Kolmogorov complexity or Solomonoff's theory of inductive inference is a promising step towards closing the neuro-symbolic gap. We would like to thank Dimitar Vasilev (Microsoft Inc.) for the computational resources used in this work.

\newpage
\small
\bibliographystyle{plainnat}
\bibliography{references}

\newpage

\appendix
\gdef\thesection{Appendix \Alph{section}}

\section{Abstraction and Reasoning Corpus}
The Abstraction and Reasoning Corpus (ARC) \citep{chollet2019measure} is a dataset of grid-based pattern recognition and pattern manipulation tasks. A decision-making agent sees a small number of examples of input and output grids that illustrate the underlying logical relationship between them. It then has to infer this logical rule and apply it correctly on a test query. 

In many ways, the benchmark is similar to the Bongard problems (view \citep{linhares2000bongard}) -- relations are highly abstract and geometric. Moreover, only 3-5 examples are presented for each task, therefore, the benchmark tests the ability of an decision-making agent to (i) grasp abstract logic and (ii) adapt quickly to new tasks. 

There are 400 training and 400 evaluation task examples, structured as follows:  
\begin{itemize}

    \item each task consists of a train and a test set;
    \item the train set includes 3-5 input/output pairs;
    \item the test set includes 1 input/output pair;
    \item an input/output pair is comprised of an input grid and an output grid, the relation between which follows a consistent logic throughout the task;
    \item grids are rectangular and are divided into $1 \times 1$ squares;
    \item grid patterns are drawn with 10 colors;
    \item grid sizes vary between 1 and 30 in length and width; input and output grid sizes are not necessarily equal.
    
\end{itemize}

No set of rules exists that can solve all tasks, and while some skills are useful for multiple of them, each task has its own unique principle. This makes trivial approaches like brute-force computation impractical.

If a human was approaching those tasks, they would easily be able to spot logical relations -- we have developed the necessary priors to find similarities and infer logic. Therefore, ideally the neural network would derive this prior during training, and that would allow it to generalize well to the evaluation dataset. 

For the current scope of our research we use all tasks with grids of size not larger than $10\times10$. For the train set, we augment the tasks to 15000 by permuting colors and by exploiting that the tasks are invariant to rotation and symmetry.

\begin{figure}[H]
    \centering
    \includegraphics[width=\linewidth]{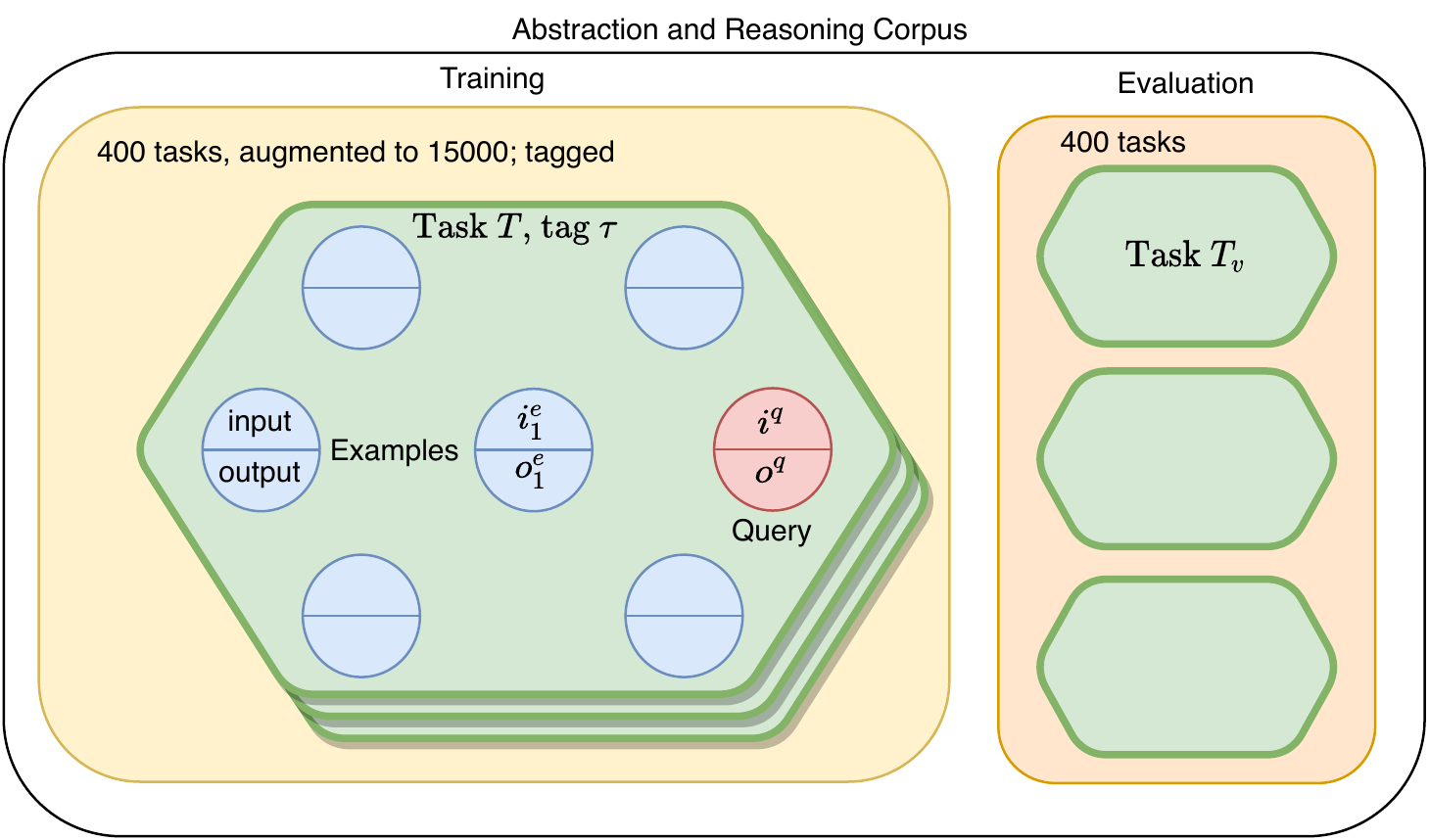}
    \caption{Structure of the Abstraction and Reasoning Corpus dataset.}
    \label{fig:arc}
\end{figure}

\newpage

\section{Grid Embedding}
\label{appa:embedding}
Prior to embedding, we zero-pad all grids to be $10\times10$, with the original grid in the center of the image. Additionally, colors in the grids are represented as one-hot vectors, making the final dimensionality of the grids $10\times 10 \times 10$ (10 colors). 

\begin{figure}
    \centering
    \includegraphics[width=\linewidth]{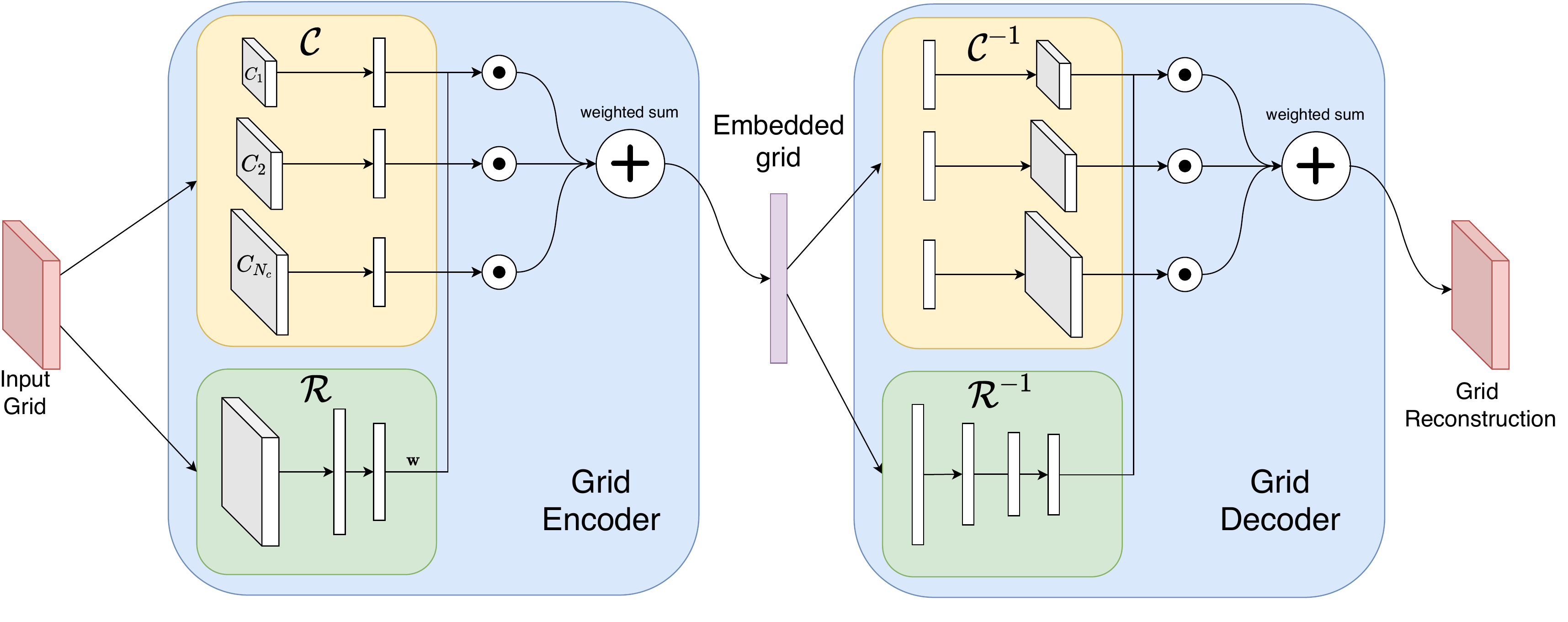}
    \caption{Structure of the Embedding network.}
    \label{fig:embedding}
\end{figure}

The embedding is done with a convolutional neural network, comprised of an encoder and a decoder. The encoder consists of a basket of 10 convolutions of filter sizes equal to $1, \dots 10$ (the $\mathcal{C}$ module, Fig. \ref{fig:embedding}). Different filter sizes enable the network to capture both local and global patterns. The convolution outputs are flattened and passed to linear layers with hyperbolic tangent activation functions, which transform them into the desired dimensionality ($n=256$). A second neural network $\mathcal{R}$ computes weights for summing the 10 resulting vectors. The decoder shares the same architecture with the encoder, but in reverse order -- first linear layers, after that convolutions and then a weighted sum; finally a softmax over the color dimension.

Summing the convolutional outputs enables the embedding network to be agnostic to the order in which it receives them (as would be with an RNN for instance). The weighs reflect the fact that grid sizes vary and therefore not all filter sizes would be equally applicable or useful.

While training the Embedding network, we found that spectral regularization again proved to be instrumental for achieving 95\% perfect reconstruction accuracy. In contrast, networks regularized by weight decay failed to climb above 6\%. 

What is more, tanh functions yielded a network that is 3 orders of magnitude more stable to Gaussian input noise than the same network, trained under the same conditions, with ReLU activations. We postulate that this is due to the fact that ReLU is unbounded for $x > 0$, therefore random perturbations would have a greater impact.

\newpage

\section{Architecture Details}
\begin{figure}[H]
    \centering
    \includegraphics{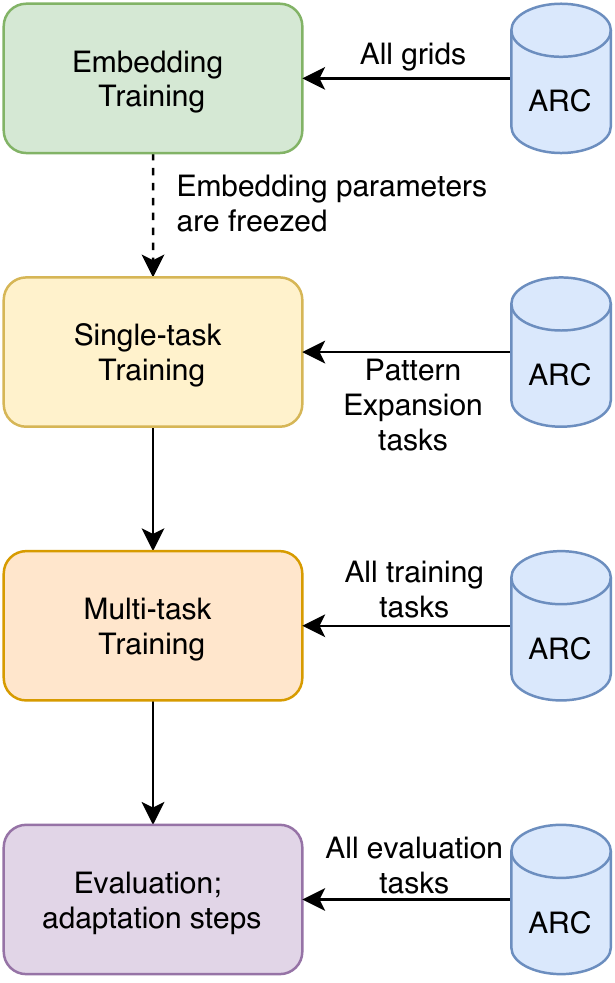}
    \caption{Order of procedures in training the model. Total runtime is $\sim 50$ hours on a single K80 GPU.}
    \label{fig:train_flow}
\end{figure}

\begin{figure}[H]
    \centering
    \includegraphics[width=0.6\linewidth]{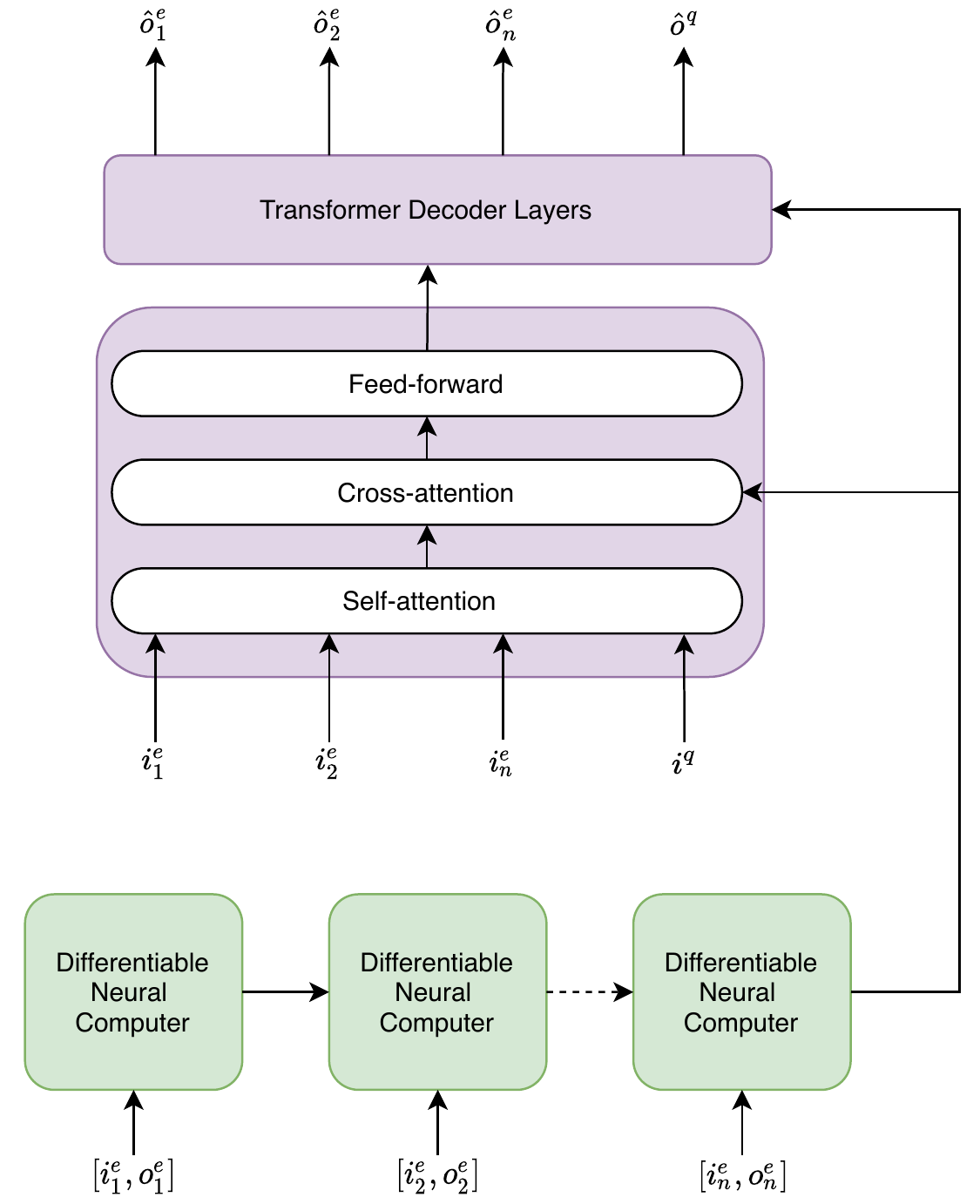}
    \caption{We use a memory-augmented neural network (DNC) to extract the context of the current task from example input-output pairs $[i_1^e, o_1^e], \dots, [i_n^e, o_n^e]$. The final DNC output is passed to the cross-attention layers of a Transformer Decoder, which processes all inputs (examples and query) and produces output predictions. Loss is calculated as by-pixel cross-entropy of the output predictions and the targets, and the model is trained end-to-end.}
    \label{fig:dnc_transformer}
\end{figure}

Figure \ref{fig:train_flow} and \ref{fig:dnc_transformer} give greater detail about the architecture we devise, as well as the procedure for training it end-to-end.
\label{appa:arch}
\vfill

\newpage

\section{Additional Results}
\begin{figure}[h]
    \centering
    \includegraphics[width=0.7\linewidth]{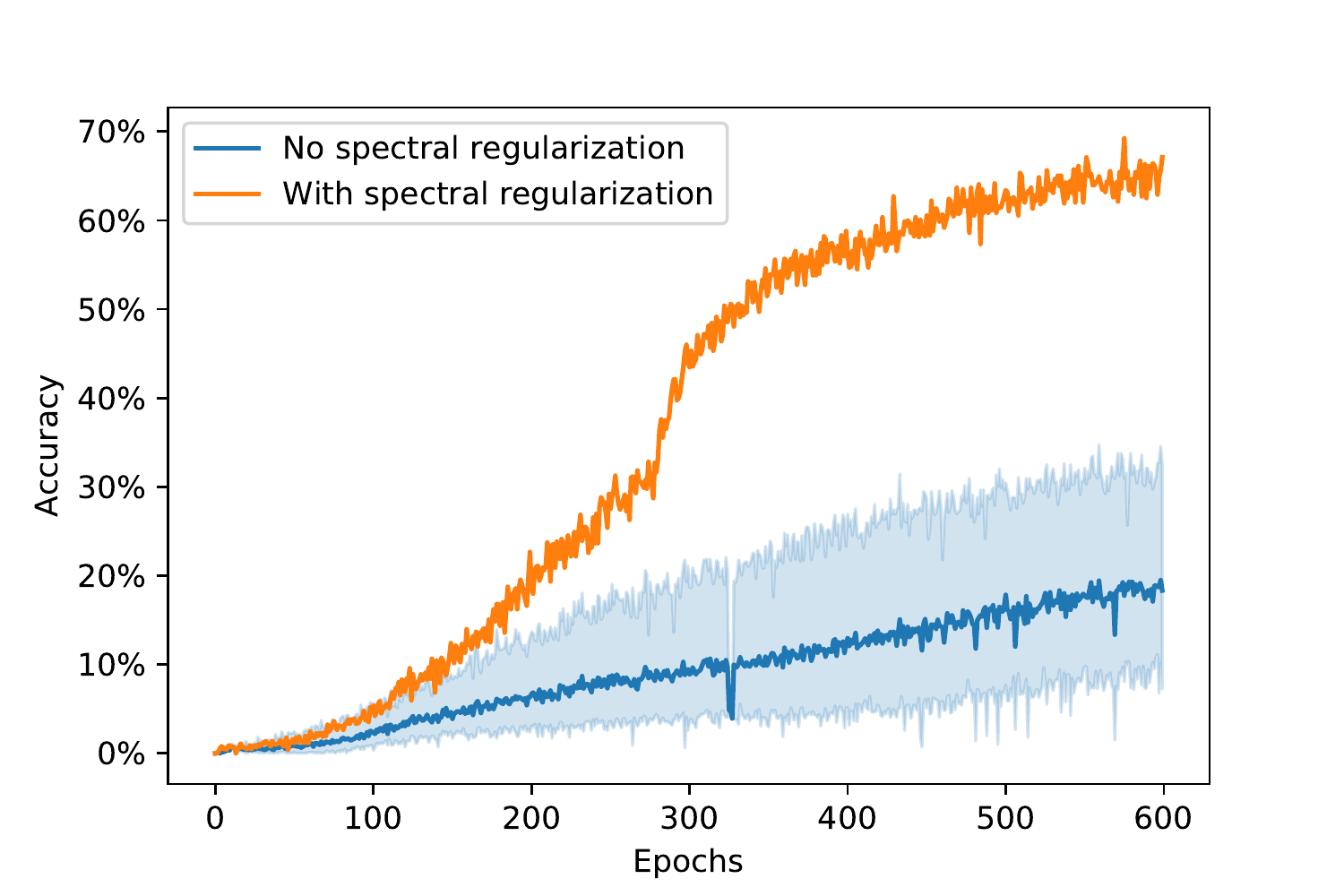}
    \caption{When we observed that stable ranks, a proxy of the number of parameters, were negatively correlated with generalization performance, we analyzed if an explicitly smaller model would improve performance without spectral regularization. While better performance and greater variance are observed, the model with spectral regularization remains unmatched. }
    \label{fig:smaller_model_results}
\end{figure}

\section{Hyperparamters}

\begin{table}[h]
\centering
\begin{tabular}{@{}lll@{}}
\toprule
Hyperparameter                  & Value          & Description                                                             \\ \midrule
learning rate                   & 0.001          & Learning rate of the models                                             \\
regularization $\lambda$        & 6e-4           & Coefficient in front of the spectral regularization penalty in the loss \\
annealing factor & 10             & Factor by which $\lambda$ is divided to weaken regularization           \\ \\
\multicolumn{3}{c}{Embedding network hyperparameters}                                                                      \\ \\
latent dimension                & 256            & The dimension of the embedded grids                                     \\
kernel sizes                    & $1, \dots, 10$ & Sizes of the used kernels in the convolutional modules                  \\
\# of kernels                   & 128            & The number of convolutional kernels                                     \\ \\
\multicolumn{3}{c}{DNC hyperparameters (temporal linkage is disabled)}                                                     \\ \\
\# of read heads                & 6              & Number of attention heads in the reading mechanism                      \\
\# of LSTM layers               & 3              & Layers of the LSTM controller                                           \\
LSTM hidden size                & 512            & Hidden size of the LSTM                                                 \\
memory dimesions                & $32 \times 64$ & Word length of 64, 32 memory locations                                  \\ \\
\multicolumn{3}{c}{Transformer hyperparameters}                                                                            \\ \\
\# of decoders                  & 4              & Number of networks in the decoder stack                                 \\
\# of attention heads           & 16             & Number of heads in the multi-head attention mechanism                   \\
Transformer hidden size         & 256            & Internal hidden size of the linear layers in the Transformer     \\
\bottomrule 
\end{tabular}

\label{tab:hyperparameters}
\end{table}

\newpage
\section{Approximation via \textit{polynomial} programs}

In this Section we briefly elaborate on the polynomial approximation mentioned in the main text. In general, classical results in this direction are based, e.g. on Chebyshev, Legendre and Bersntein polynomial approximations. Here we discuss a 1-dimensional approximation via Bernstein polynomials, but similar estimates hold in higher dimensions as well as other polynomial schemes (e.g. Chebyshev). We refer to \cite{Canuto2006, Lloyd2013} for a thorough collection of results as well as further references.

A well-known method of approximating one-dimensional continuous functions is by means of Bernstein polynomials. Let $f:[0, 1] \rightarrow \mathbf{R}$ be continuous. The Bernstein polynomial $B_n(f; t)$ of order $n$ corresponding to $f$ is defined as:

\begin{equation}
     B_n(f; t) := \sum_{k = 0}^n f\left( \frac{k}{n} \right) \binom{n}{k} t^k (1-t)^{n-k}.
\end{equation}{}

\begin{theorem} \label{thm:Bernstein}
     The following estimate holds:
     \begin{equation}
         \|f- B_n(f; t) \|_{L^\infty} \leq \frac{3}{2} \omega \left( f; \frac{1}{\sqrt{n}} \right),
     \end{equation}
     where $\omega$ denotes the modulus of continuity defined as
     \begin{equation}
         \omega(f; \delta) := \sup \left\{ |f(x) - f(y)| : x, y \in [a, b], |x - y| \leq \delta  \right\}.
     \end{equation}
     In other words, $\omega$ measures the maximal jump $|f(x) - f(y)|$ over points $x, y$ which are no more than a distance $\delta$ apart.
\end{theorem}{}

Note that whenever $f$ has a Lipschitz constant $L$, the modulus of continuity $\omega(f, \delta)$ is controlled above via $L\delta$. Applying this bound in the estimate of Theorem \ref{thm:Bernstein} yields the result mentioned in the text. As already mentioned, more technical high dimensional analogues of the estimates are also available \cite{Lloyd2013, Canuto2006} - e.g. an analogues result for Chebyshev polynomials holds where the modulus of continuity $\omega$ is appropriately replaced by extrema over complex ellipsoids \cite{trefethen2017a}.

\end{document}